\documentclass[letterpaper]{article}
\usepackage{preprintstyle}
\usepackage[hyphens]{url}
\usepackage{graphicx}
\usepackage{natbib}
\usepackage{caption}
\usepackage{algorithm}
\usepackage{algorithmic}

\usepackage{newfloat}
\usepackage[utf8]{inputenc}
\usepackage{tabularx}
\usepackage{enumitem}
\usepackage{amsmath}
\usepackage{amssymb}
\usepackage{listings}
\usepackage{array}
\usepackage[table]{xcolor}

\usepackage{booktabs}
\usepackage{pifont}
\usepackage{xcolor}
\usepackage[utf8]{inputenc}
\usepackage{textgreek}
\usepackage{booktabs}
  \usepackage{xcolor}
  \usepackage{colortbl}        

  \definecolor{ftrow}{gray}{0.93}

\newcommand{\gain}[1]{\textcolor{teal}{#1}}
\newcommand{\loss}[1]{\textcolor{red!70!black}{#1}}

\DeclareCaptionStyle{ruled}{labelfont=normalfont,labelsep=colon,strut=off} 
\floatstyle{ruled}
\newfloat{listing}{tb}{lst}{}
\floatname{listing}{Listing}

\usepackage{booktabs}

\title{DementiaCare-Bench: A Modality-Validated Video Benchmark
}
\author{
    Afrouz Sheikholeslami\textsuperscript{\rm 1},
    Yuankai Qi\textsuperscript{\rm 1}\corresponding,
    Xuyun Zhang\textsuperscript{\rm 1},
    Luping Zhou\textsuperscript{\rm 2},\\
    Amin Beheshti\textsuperscript{\rm 1},
    Quan Z. Sheng\textsuperscript{\rm 1},
    Ming-Hsuan Yang\textsuperscript{\rm 3}
}
\affiliations{
    \textsuperscript{\rm 1}School of Computing, Macquarie University, Sydney, NSW, Australia\\
    \textsuperscript{\rm 2}School of Electrical and Computer Engineering, University of Sydney, Sydney, NSW, Australia\\
    \textsuperscript{\rm 3}University of California, Merced, CA, USA\\
    afrouz.sheikholeslami@hdr.mq.edu.au, yuankai.qi@mq.edu.au
}

\begin{document}

\maketitle

\begin{abstract}
Dementia affects an estimated 57 million people worldwide, and for most families the hardest part of care is not memory loss but the behavioral and psychological symptoms of dementia (BPSD): agitation, wandering, resistance to care, sundowning. 
Understanding these symptoms requires more than recognizing the behavior itself; it also requires knowing what happened beforehand. The same behavior may call for a different response depending on its trigger.
Video-language models (VLMs) could potentially support caregivers, yet no existing benchmark evaluates this capability.
To fill this gap, we present DementiaCare-Bench: 56 professionally produced caregiver-training videos segmented into 94 clips across nine BPSD categories, with 2,023 questions generated by a multi-agent pipeline that grounds every clinical claim in a verbatim transcript span. Each question is then probed under four visual conditions and labelled by the least it requires, so its visual demand is measured rather than assumed. 
Measurement contradicts intent: we wrote 77.7\% of the questions to require ordered frames, and 34.8\% do. 
Across 12 current VLMs the pattern is uniform. The best reach 85\% overall, but that average is carried by questions a language model can answer from clinical knowledge alone; accuracy falls by 17 points on average on questions that require the ordered clip, and a leading open model scores at chance on judging whether a caregiver's response was appropriate. A lightweight LoRA fine-tune, DemCare-VLM, moves video dependence from $-3.3$ to $+4.5$ points, so what the benchmark exposes can be repaired and not only measured.
\end{abstract}

\section{Introduction}
 Most dementia care is delivered at home, by family members with no clinical training \cite{friedman2015informal}. What makes it hard is rarely memory loss. It is the behavioral and psychological symptoms of dementia (BPSD): agitation and aggression, wandering and exit-seeking, resistance to daily care, and the disorientation of sundowning. BPSD are the leading driver of caregiver burden and burnout and a primary determinant of the decision to move a person with dementia into institutional care \cite{kales2015bpsd}. An estimated 57 million people worldwide live with dementia, a number projected to nearly triple by 2050 as populations age, and the population of untrained caregivers facing these symptoms will grow with it \cite{gbd_dementia_forecasting_2022}.


Video-language models (VLMs) offer a plausible form of support: a
system that could interpret a caregiving interaction might provide
real-time guidance, interactive training, or monitoring \citep{abid2026vlmhealthcare}. Interpreting these interactions requires understanding how events unfold in time. 
%
Consider a person with dementia pushing the caregiver's hand away
during dressing. If the push follows the caregiver reaching for their
arm without warning, it is a startle response, and the caregiver
should step back into the person's field of view and announce contact
before touching. If the same push arrives unprompted, it more likely
signals pain or discomfort, and the caregiver should stop the task and
check for a physical cause rather than adjust their approach. The
action on screen is identical in both cases. What separates them is
the event immediately preceding it, which identifies the trigger, and
the trigger is the one part of the situation a caregiver can change
\citep{kales2015bpsd}.
However, no benchmark measures this capability for dementia care. Adjacent work is limited to
sensor-based agitation detection, which registers that a behavior is
occurring without interpreting it \citep{sensoragitation}, and to
caregiving LLMs evaluated on text alone \citep{textllmdementia}.

Building such a benchmark faces a second problem, one not specific
to dementia. VideoQA benchmarks cannot establish that their own
questions require video: a question phrased around event order may be
answerable from language priors, from a single frame, or from
co-occurrence statistics that bypass the clip entirely
\citep{cheng2025vstar}. 
The consequence is that scores stop being
interpretable. 
We therefore measure, for every question, how much of the clip
answering it requires. Each candidate goes to a fixed reference model
under four conditions: no frames, a single frame, all frames shuffled,
and all frames in order. The question takes the label of the least
visual information that suffices. A question answered from shuffled
frames is labelled Semantic rather than Temporal, because a model that
succeeds on scrambled frames has demonstrated that it does not use
their order. Validation runs inside the generation pipeline, so every
released question carries a label established before publication
rather than audited after it.

Building on this modality validation, we introduce \textbf{DementiaCare-Bench}, a modality-validated VideoQA benchmark for dementia caregiving. It comprises 56 caregiving videos
from 11 publicly available training sources, segmented into 94 clips
across nine clinically grounded BPSD categories, with 2,023
multiple-choice questions in 14 categories. 38 of the videos are
paired: the same incident appears twice, once with the common
caregiver response and once with the expert-recommended one. A
question about whether a response was appropriate therefore has
identical text across the two clips and no answer available from the
question alone. Questions are produced by a multi-agent pipeline in
which every clinical claim is grounded in a verbatim transcript span,
and each carries a verified modality label with its per-condition
evidence.

 
Our main contributions are:
\begin{enumerate}
\item \textbf{DementiaCare-Bench}. A video-language benchmark
for dementia caregiving: 94 clips across nine BPSD categories, with a question set spanning perception, temporal, causal, and clinical-reasoning capabilities. Reading a BPSD episode means tracking what happened, in what order, and what set it off, so the benchmark tests each step separately and reports where a model's competence stops.
\item \textbf{Modality Validation}. Certifies per question whether answering requires ordered video, a single frame, or text alone, inside the generation loop rather than as a post-hoc audit. Applied to our own generators it shows that 77.7\% of questions written to require ordered video reduce to 34.8\% once measured. To our knowledge this is the first quantification of the error a benchmark incurs by assigning capability labels from design intent.

\item \textbf{DemCare-VLM}. A model trained to condition on the clip rather than on the question. Every question appears in training twice, once with frames from its own clip and once with frames from an
unrelated clip, and the model is penalised for answering when the
frames do not support the question. Video dependence moves from
$-3.3$ to $+4.5$ points.
\end{enumerate}

\section{Related Work}
\subsection{Video-Language Models}
Modern video-language models couple a visual encoder with a
pretrained LLM~\citep{videochatgpt, llavaonevision, qwen3vl,
internvl3, gpt4o, gemini25}, with a dedicated line of work on
temporal modeling: recurrent memory~\citep{videollamb},
hierarchical token compression~\citep{internvideo25}, and
principled training recipes~\citep{apollo}. Yet temporal reasoning
consistently lags static perception, and most models ingest
uniformly sampled frames regardless of where the answer-bearing
moments lie~\citep{vbenchcomp, egotempo}. DemCare-VLM builds on
this observation, adapting a strong open-weight model through
lightweight fine-tuning rather than architectural change.

\subsection{Video Understanding Benchmarks}
VideoQA benchmarks have grown from short-clip perception
tests~\citep{nextqa, mvbench} to long-video
suites~\citep{videomme, egoschema, perceptiontest} and
expert-level material spanning professional
lectures~\citep{videommmu}, skilled physical
activity~\citep{exact}, and social interaction~\citep{sivbench}.
A diagnostic literature, however, shows that aggregate scores on
such benchmarks are inflated by shortcuts: text-only models exceed
chance by exploiting language priors~\citep{lostintime, mvp}, and
accuracy is often invariant to frame shuffling, indicating that
nominally temporal questions are answerable from static
cues~\citep{vbenchcomp, egotempo}. 


Domain-specific video benchmarks have begun to appear in medicine.
DrVD-Bench structures questions around stepwise clinical reasoning
\citep{drvdbench}, but over diagnostic imaging rather than human
interaction. Computational work on dementia has centred on detection:
wearable and ambient sensing identifies that agitation is occurring
\citep{sensoragitation} without any language understanding, while LLM
applications evaluate caregiver support from text alone
\citep{textllmdementia}. Between these lies the capability a caregiving
assistant needs and no benchmark measures: reading a temporally
extended interaction in which the clinical meaning of a behaviour
depends on the order and cause of events. DementiaCare-Bench fills
that gap, and its paired clips additionally test whether a model can
judge response appropriateness, the judgment at the heart of caregiver
training \citep{kales2015bpsd}.


\section{DementiaCare-Bench Construction}
 
\subsection{Data Collection and Clip Structure}

DementiaCare-Bench is built from 56 dementia-caregiving videos drawn from
11 publicly available caregiver-training programs and educational channels.
The sources are professionally produced training material in which the
people with dementia are portrayed by actors, not real patients;
the videos are designed to teach family and professional caregivers how to
recognize and respond to behavioral symptoms. The full list of sources,
with provenance and links, is provided in the supplementary material. We organise the corpus around nine categories of
behavioral and psychological symptoms of dementia (BPSD): Agitation \& Aggression, Wandering \& Exit-Seeking, Apathy \& Withdrawal, Disinhibition, Repetitive Behavior, Delusions \& Hallucinations, Care Resistance (resistance during activities of daily living), Sleep Disturbances \& Sundowning, and Other. These categories adapt the symptom domains of the Neuropsychiatric Inventory \citep{TBD-npi} and Cerejeira et al.'s clinical review of BPSD \citep{TBD-bpsd-review}.

\paragraph{Video structure.} Most training videos follow a three-part structure built around a single behavioral incident. A caregiver first demonstrates the common response, instinctive but typically counter-productive. An expert then explains what drives the behavior and why that response fails. The video closes with the recommended response.
Of the 56 source videos, 38 follow this structure and yield a common-response clip and a recommended-response clip each (76 clips). The remaining 18 lack a clean common-versus-recommended contrast and are retained as single full-length clips, giving 94 clips total. Since the two clips share a behavioral situation but differ only in the caregiver's response, the benchmark can pose comparative questions that require a model to judge whether a response is appropriate. We treat common-response and recommended-response understanding as separable axes and report them separately. The expert explanation is not extracted as a clip but grounds question generation.

\begin{figure}[t]
\centering
\includegraphics[width=\columnwidth]{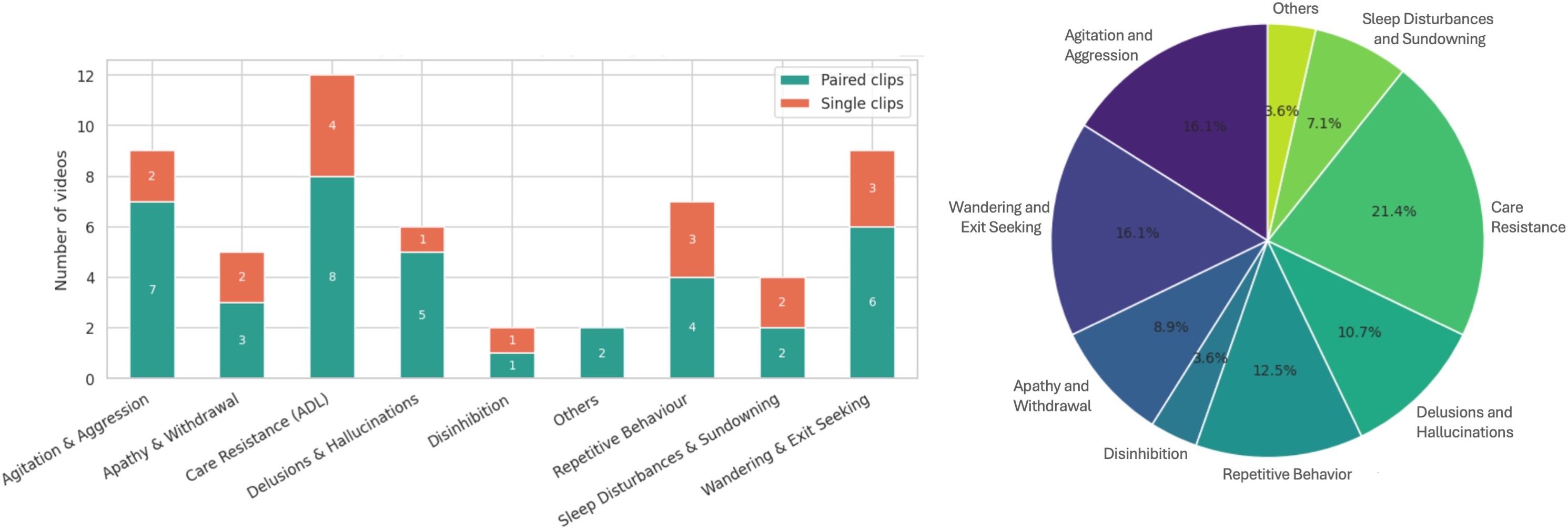}
\caption{Left: number of videos per BPSD category, split by clip type (paired vs.\ single). Right: proportion of videos per category across the full corpus. Care Resistance, Agitation \& Aggression, and Wandering \& Exit-Seeking dominate; the Others category consists entirely of single clips.}
\label{fig:corpus-stats}
\end{figure}

The 94 clips span 4.7 hours, with a mean duration of 3.0 minutes. Figure~\ref{fig:corpus-stats} shows the distribution
across BPSD categories. Care Resistance, Agitation \& Aggression, and
Wandering \& Exit-Seeking are the best represented and the most
pairing-complete. Full breakdowns by source and duration are in the supplementary material.

\subsection{Question Categories and Generation Pipeline}
\label{subsec:QA_generation}

Caregiver training for BPSD is organized around a clinical reasoning process. Most influentially the DICE approach \cite{kales2015bpsd} describes the behavior
and its context, investigates what triggered it, creates a response, and evaluates
whether it worked. These steps specify what an assistant supporting a caregiver
has to do, and we adapt them into the four reasoning axes of DementiaCare-Bench:
visual perception, temporal and causal reasoning, clinical reasoning, and comparative analysis. Each axis subdivides into
fine-grained categories, fourteen in total. Table~\ref{tab:categories}
defines every category with a representative question.

\paragraph{Visual perception.} Before any interpretation, a caregiver must see the scene accurately. This axis comprises four subcategories, including the person's action, caregiver's action, scene and environment, and person's appearance. Every answer is readable directly from the video, establishing a perceptual floor for the axes that follow.

\paragraph{Temporal and causal reasoning.} Dementia behaviors unfold over time and have a structure: a trigger, a response, an escalation or resolution. This axis covers five subcategories: action count, action order, causal attribution, caregiver's and person's action sequence. Causal attribution carries the most clinical weight. Caregiver training centers on identifying the trigger because it is the one part of the situation a caregiver can change.

\paragraph{Clinical reasoning.} Beyond observing the scene, a caregiver must interpret it: name the behavioral symptom, judge its severity and escalation risk, track how the person's emotional state shifts across the clip, and recommend a safe response. Unlike the previous axes, correct answers here require domain knowledge, knowing not just what is happening but what it means clinically and what the appropriate response is.

\paragraph{Comparative analysis.} The paired clips enable an axis single-clip datasets cannot support. Both clips show the same incident, but the caregiver responds differently: once with the common instinctive response, and once with the expert-recommended one. A model must contrast the two and judge why one is safer or more effective. 



\begin{figure*}[t]
\centering
\includegraphics[width=\linewidth]{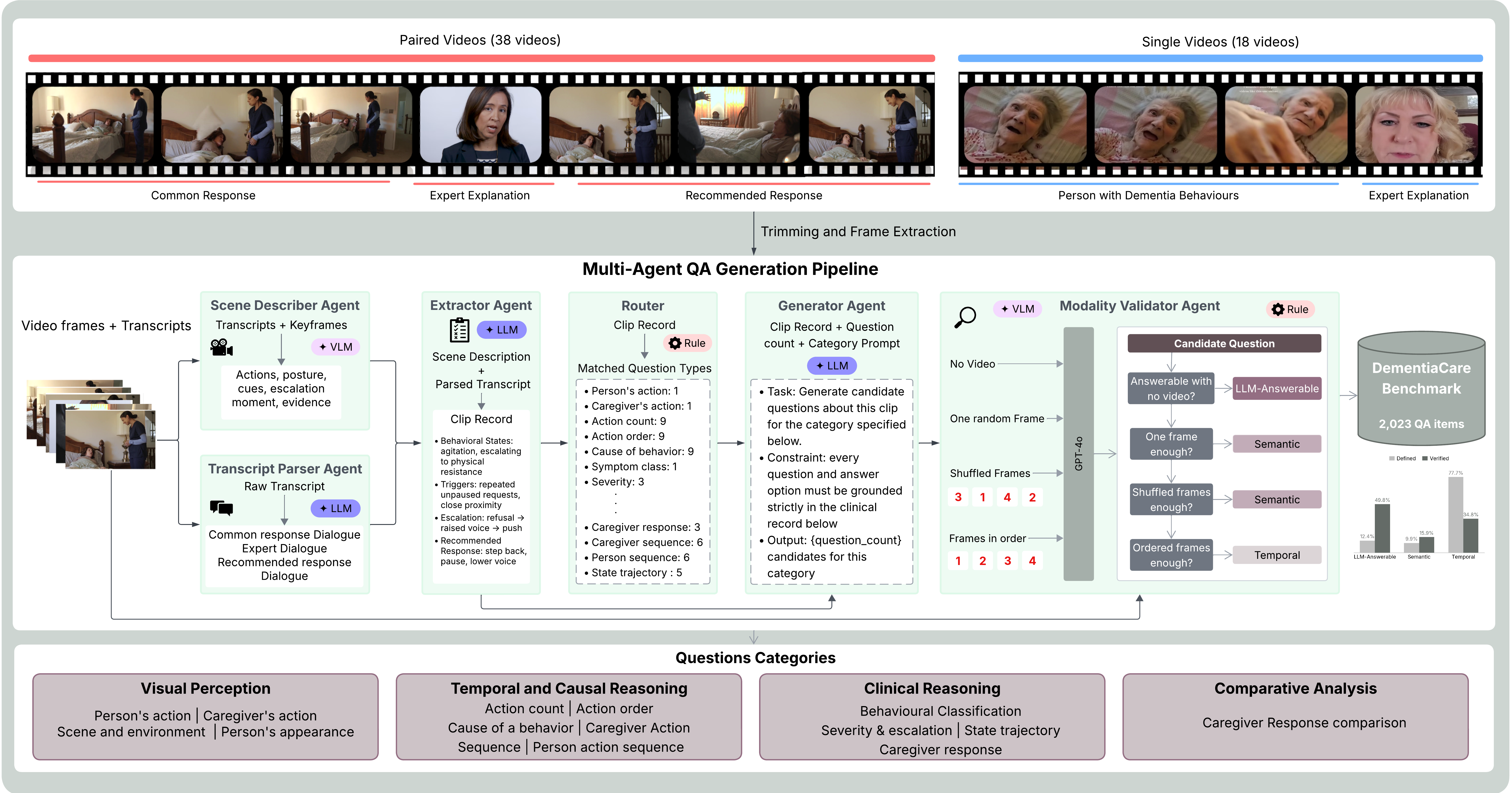}
\caption{The multi-agent question generation and validation pipeline. 
}
\label{fig:pipeline}
\end{figure*}

\subsubsection{Generation pipeline.} 
We generate and validate questions
with a multi-agent pipeline, governed by one constraint: every
question must be grounded in what the clip shows or in the expert
commentary. Each agent is a prompted instance of a general-purpose
model rather than a trained component. Qwen3-VL-235B-A22B backs the
agents that describe scenes, parse transcripts, and write questions;
GPT-4o backs the Modality Validator. We deliberately validate with a
different model from the one that generates, so that a question is not
certified by the model that wrote it. Full prompts are in the
supplementary material. Each agent handles one stage and passes its
output to the next. Concretely, the pipeline runs in five stages, each handled by a different agent that passes its output to the next:

\begin{enumerate}[label=(\roman*),leftmargin=*]
\item \textbf{Scene description.} A vision agent processes up to twelve keyframes and produces a structured account of the visible scene, such as actions, posture, proximity, facial and vocal cues.
\item \textbf{Transcript parsing.} A parser splits the transcript into scene dialogue and expert narration. This isolates what is said during the incident from what the expert teaches about it.
\item \textbf{Extraction.} An extractor fuses the scene description and parsed transcript into a clinical record per clip: the behavioral state, trigger, escalation pathway, recommended response, caregiver errors, and safety constraints. Every clinical claim in the record must be traceable to a specific line in the transcript. If the transcript does not support a field, that field is excluded. Generators then write questions only from this record.
\item \textbf{Routing and Generation.} Not every clip supports every question category. A clip with no repeated actions cannot yield action-count questions, and a clip with no clear trigger cannot yield causal attribution questions. A router inspects each clip's record and sets a quota per category based on what the clip can support, then dispatches each quota to a category-specific generator. For temporal categories, the router over-generates candidates.

\item \textbf{Modality validation.} Each candidate is probed under four visual conditions (Section Modality Validation). Items that pass receive a verified-modality label; the rest are discarded. Only validated questions enter the benchmark.
\end{enumerate}

\begin{table}[t]
\centering
\small
\caption{The four reasoning axes of DementiaCare-Bench, with one
representative question each and the verified modality of that
question. All fourteen fine-grained categories, with an example for
each, are listed in the supplementary material.}
\label{tab:categories}
\setlength{\tabcolsep}{4.5pt}
\renewcommand{\arraystretch}{1.2}
\begin{tabular}{@{}>{\scriptsize\raggedright\arraybackslash}p{0.17\columnwidth}
                  >{\scriptsize\raggedright\arraybackslash}p{0.57\columnwidth}
                  >{\scriptsize\raggedright\arraybackslash}p{0.14\columnwidth}@{}}
\toprule
\textbf{Axis} & \textbf{Representative question} & \textbf{Modality} \\
\midrule
\rowcolor{black!6}
Visual perception \newline (4 categories) &
What does the caregiver do with their hands when the person with
dementia becomes physically aggressive during the dressing routine? &
Semantic \\
Temporal \& causal reasoning \newline (5 categories) &
Which environmental trigger or caregiver action immediately preceded
the person with dementia's verbal outburst? &
Temporal \\
\rowcolor{black!6}
Clinical reasoning \newline (4 categories) &
What should the caregiver do to encourage the person with dementia to
take their medication without using physical force or arguing? &
LLM-Answerable \\
Comparative analysis \newline (1 category) &
How does the caregiver's approach differ between the clip where they
argue with the person's delusion and the clip where they use
validation? &
Temporal \\
\bottomrule
\end{tabular}
\end{table}

\subsection{Modality Validation}
\label{sec:verification}
 
A question tests video understanding only if answering it requires the
video. We measure that requirement for every question rather than
inferring it from the question's wording.
 
Each candidate goes to a fixed reference model, GPT-4o,
under four conditions: \textbf{no frames}, the question and its options
alone; \textbf{one frame}, a single sampled frame; \textbf{shuffled
frames}, all sampled frames in random order; and
\textbf{ordered frames}, the same frames in temporal order. Each
condition runs 5 times with the answer options reordered, and
passes only when every run is correct. A question then takes the label
of the least visual information that suffices:
 
\begin{itemize}[leftmargin=*,itemsep=0pt,topsep=2pt,parsep=0pt]
\item \textbf{LLM-Answerable.} Correct with no frames. We retain these
as a language-prior control and report them apart from the rest of the
benchmark.
\item \textbf{Semantic.} Correct from one frame or from shuffled
frames, wrong with none. Answering needs visual evidence but not the
order of events.
\item \textbf{Temporal.} Correct only from ordered frames.
\end{itemize}
 
The shuffled condition is what gives the Temporal label its content. A
model that answers correctly once the frames are scrambled does not
need their order, whatever the question's phrasing implies, so we label
it Semantic. Candidates that no condition renders answerable are
discarded, so every question in DementiaCare-Bench is one the reference
model answers given the ordered clip: a model scoring poorly on any
subset is failing questions already shown to be answerable.
 

\subsection{Benchmark Statistics}
DementiaCare-Bench contains 2{,}023 questions over 94 clips. Each
question is stored as a self-contained record carrying its category,
its verified modality, the per-condition outcomes backing that
modality, the question format, the question and answer, and references
to the ground-truth fields it draws from.
\paragraph{Defined versus verified modality.}
Every question carries two labels: the modality its generator was
written to produce, and the modality validation certifies. They
disagree sharply. We wrote 77.7\% of the questions to require ordered
frames and 12.4\% as language-prior controls; measurement returns
34.8\% and 49.8\%. Semantic items move least, from 9.9\% to 15.9\%.
Reported under defined labels, as VideoQA benchmarks conventionally
are, DementiaCare-Bench would have overstated its temporal content by
a factor of 2.2. Our generators write from a clinical record extracted
from the transcript, and a question written from a text description of
a sequence tends to be answerable from text. We measure this for our
own generators and cannot measure it for benchmarks that do not
publish defined labels; the point is that the gap is large enough to
matter and cheap enough to check.


\section{Method: DemCare-VLM}
DemCare-VLM adapts Qwen3-VL-32B to dementia-care video with one objective: make the model read the clip before it answers, rather than guess from the question text. We keep the base model frozen, train a small LoRA adapter, and build the training data so that a model leaning on language priors loses points on half of every batch.

\subsection{LoRA fine-tuning}
We freeze all the pretrained parameters and insert low-rank
adapters \cite{Hu2021LoRALA} into the four attention projections of every self-attention block.
We leave the vision encoder and the feed-forward layers untouched,
because attention is where the model weighs visual tokens against text,
and therefore where the shortcut lives.
The adapter uses rank $r{=}16$ and scaling $\alpha{=}32$, adding
39.8\,M trainable parameters (0.12\% of the model).
We train for three epochs with AdamW at a peak learning rate of
$10^{-4}$, cosine decay after a 5\% linear warmup, and an effective
batch size of 8 via gradient accumulation.
Because the base weights never change, each checkpoint stores only the
adapter tensors (a few hundred megabytes), and a full run fits on four
consumer GPUs.
\subsection{Training data}
We build examples from the multiple-choice questions on the 38 paired
videos, split at the video level: 31 videos for training and 7 for
testing, so no clip appears on both sides of the boundary.
Each training example takes one of two forms.

A \textbf{matched} example pairs the model with six evenly-spaced
frames from the clip the question was written for, the question text,
and its answer options; the target is the correct letter.
These examples teach the benchmark vocabulary and the mapping from a
specific clip to its answer.

A \textbf{mismatch negative} keeps the same question but substitutes
frames from a random clip drawn from a different BPSD category.
The target becomes Insufficient visual evidence to answer
this question.
A model that answers from question text alone produces the letter it
would have chosen for the matched case and scores the negative wrong;
to answer correctly, the model must check whether the frames support
the question and abstain when they do not.
We add one negative per matched example, giving a 50/50 split in the
training set.
In both cases we compute loss on the answer tokens only, masking the
question and frame tokens with the ignore index ($-100$), so the
adapter learns to select the answer rather than echo the prompt.
The final training set contains 1,940 examples; the test set contains
297 matched examples.

\subsection{Relation to prior work}
LoRA fine-tuning and answer-only loss masking are standard practice
\cite{Hu2021LoRALA}.
Our contribution is the reject-on-mismatch
negative, which converts the benchmark's own diagnosis into a training
signal: the language-prior shortcut quantified above becomes the
behaviour the negatives penalise. Rather than teaching the model more
about dementia care, it teaches the model to check whether the frames
in front of it bear on the question, and to abstain when they do not.
For a caregiving assistant, that abstention is not a failure mode but
the correct response to insufficient observation.

\begin{table*}[!t]
  \centering
  \small
  \caption{Accuracy (\%) on DementiaCare-Bench (all 2,023 questions). \textbf{Left block}: overall accuracy at each evaluated frame count; 0 denotes text-only prompting. \textbf{Summary}: Best is the highest accuracy over the frame-supplied settings only (8--128);
           $\Delta_v$ is the Best minus the text-only score. \textbf{Right block}: accuracy at each model's Best frame count, broken down by verified modality. 'LLM-ans.' is answerable from text alone; 'Sem.' requires visual evidence but not frame order; 'Temp.' requires the ordered frame sequence.}
  \label{tab:ablation}
  \renewcommand{\arraystretch}{1.05}
  \setlength{\tabcolsep}{6pt}
  \begin{tabular}{l ccccccc |cc| ccc}
  \toprule
  & \multicolumn{7}{c}{\textbf{Overall accuracy by frame count (\%)}}
  & \multicolumn{2}{|c|}{\textbf{Summary}}
  & \multicolumn{3}{c}{\textbf{At Best, by modality (\%)}} \\
  \cmidrule(lr){2-8}\cmidrule(lr){9-10}\cmidrule(lr){11-13}
  \textbf{Model}
    & \textbf{0} & \textbf{8} & \textbf{16} & \textbf{32}
    & \textbf{48} & \textbf{64} & \textbf{128}
    & \textbf{Best} & $\boldsymbol{\Delta_v}$
    & \textbf{LLM-ans.} & \textbf{Sem.} & \textbf{Temp.} \\
  \midrule
  \multicolumn{13}{l}{\textit{Proprietary}} \\
  Claude-Sonnet-4-6
    & 83.7 & 81.2 & 83.7 & 82.8 & 83.2 & 83.2 & 83.1
    & 83.7 & $\phantom{+}$0.0
    & 87.3 & 85.7 & 69.8 \\
  Gemini-2.5-Flash
    & 81.0 & 81.9 & 81.9 & 81.0 & 80.5 & 81.4 & 80.8
    & 81.9 & $+$0.9
    & 85.9 & 84.9 & 65.4 \\
  GPT-4o
    & 80.1 & 77.4 & 78.7 & 78.8 & 78.8 & 78.3 & 78.6
    & 78.8 & $-$1.3
    & 82.8 & 83.2 & 61.8 \\
  \midrule
  \multicolumn{13}{l}{\textit{Open-source}} \\
  Gemma-3-27B-IT
    & 84.8 & 83.5 & 84.1 & 84.7 & 84.9 & 85.1 & 85.1
    & 85.1 & $+$0.3
    & 88.2 & 87.1 & 67.4 \\
  Qwen3-VL-32B
    & 83.6 & 77.6 & 80.5 & 81.8 & 80.5 & 80.4 & 80.4
    & 81.8 & $-$1.8
    & 91.9 & 68.5 & 65.4 \\
  Qwen3-VL-235B-A22B
    & 82.3 & 81.0 & 81.0 & 82.0 & 82.1 & 81.3 & 82.0
    & 82.1 & $-$0.2
    & 85.7 & 87.8 & 65.1 \\
  Qwen3-VL-30B-A3B
    & 80.5 & 79.9 & 79.2 & 79.9 & 78.3 & 78.3 & 78.7
    & 79.9 & $-$0.6
    & 85.2 & 80.3 & 61.0 \\
  Qwen3-VL-8B
    & 78.6 & 75.2 & 79.1 & 79.4 & 78.7 & 78.6 & 79.4
    & 79.4 & $+$0.8
    & 82.5 & 85.3 & 64.6 \\
  GLM-4.6V
    & 73.2 & 71.9 & 70.8 & 71.2 & 72.5 & 71.2 & 71.1
    & 72.5 & $-$0.7
    & 78.1 & 77.1 & 46.5 \\
  InternVL3-8B
    & 70.4 & 63.5 & 65.9 & 64.7 & 65.8 & 65.8 & 65.7
    & 65.9 & $-$4.5
    & 69.7 & 71.3 & 48.6 \\
  Keye-VL-8B
    & 62.5 & 73.1 & 75.0 & 75.0 & 74.8 & 74.7 & 74.7
    & 75.0 & $+$12.5
    & 77.5 & 79.6 & 62.8 \\
  MiniCPM-V-2.6
    & 53.2 & 56.5 & 58.1 & 59.0 & 59.9 & 59.4 & 58.9
    & 59.9 & $+$6.7
    & 63.6 & 65.9 & 42.4 \\
  \bottomrule
  \end{tabular}
  \end{table*}
\section{Experiments}
\label{sec:experiments}



\paragraph{Evaluated Models.}
We evaluate twelve VLMs spanning proprietary and open-weight families
(Table~\ref{tab:ablation}).
Proprietary models, including GPT-4o, Gemini-2.5-Flash, and Claude-Sonnet-4-6 which are
accessed via API\@.
Open-weight models span two architectural classes: MoE models
(Qwen3-VL-235B-A22B, Qwen3-VL-30B-A3B) and dense models at the 27--32B scale
(Gemma-3-27B-IT, Qwen3-VL-32B, GLM-4.6V) and the efficient 8B scale
(Qwen3-VL-8B, InternVL3-8B, Keye-VL-8B, MiniCPM-V-2.6). We report video dependence as $\Delta_v$, the difference between a model's best accuracy with frames and its accuracy with none. A negative $\Delta_v$ means the model does better without the video.

\paragraph{Frame-count ablation.}
All models are evaluated under a frame-count ablation, where each question is presented with \(\{0, 8, 16, 32, 48, 64, 128\}\) sampled frames. The case of 0 frames denotes text-only prompting, using the question stem and answer choices without visual input. For video inputs, frames are extracted at 1 fps and then uniformly sampled at equal temporal intervals from each clip. All frames are resized to \(512\times512\) for consistency.


\subsection{The Language-Prior Shortcut}
\label{sec:shortcut}
 
Prior work has shown that video question-answering benchmarks are partially
solvable from language priors: text-only language models exceed chance on
MVBench and NExT-QA by exploiting cues in the question and answer options \cite{lostintime,mvp,sivbench}.
On DementiaCare-Bench, this effect is not partial but near-total for the
strongest models.
Table~\ref{tab:ablation} reports each model's accuracy without any frames
and at its best-performing frame count, together with the video-dependence
gain, which we define as the difference between the two.
 

Six of the twelve models score strictly highest with no frames at all (Table~\ref{tab:ablation}), and a seventh, Claude-Sonnet-4-6, ties its best frame-supplied score. The pattern is sharpest at the top of the table. Ranked by text-only accuracy, the four leading models are Gemma-3-27B-IT (84.8), Claude-Sonnet-4-6 (83.7), Qwen3-VL-32B (83.6) and Qwen3-VL-235B-A22B (82.3), and for none of them do frames buy anything measurable: their video-dependence gains are $+0.3$, $0.0$, $-1.8$ and $-0.2$.

Aggregate accuracy conceals where the shortcut operates. Broken out by
question category, the text-only accuracy of Qwen3-VL-32B descends in
three steps. On clinical safety questions, which ask what a caregiver
should do in principle and are answerable from medical facts encoded
during pretraining, it is correct every time: 100\% without a single
frame. On person-attribute questions it reaches 94\%, because the
question stem names the attribute it asks about. On questions asking
whether the caregiver's response shown in the clip was appropriate, a
two-way choice that cannot be answered without watching what the
caregiver did, it scores 50\%, which is chance. The model has
memorised the caregiver-training curriculum and cannot recognise
whether the caregiver in front of it followed it.

 
These results show that a model with weaker prior knowledge should
rely on the video more, and the results bear this out.
The two models that gain the most from frames, Keye-VL-8B and MiniCPM-V-2.6,
are also the weakest in the text-only setting, at 62.5 and 53.2 percent
respectively; supplying frames raises their accuracy by 12.5 and 6.7 points.
Their improvement is not evidence that they understand the clips better than
the stronger models, but that their thinner priors leave more room for
visual evidence to contribute.
Rather than undercutting the shortcut account, these two models serve as a
control that supports it.
 

 
\subsection{DemCare-VLM Restores Video-Dependence}
\label{sec:demcarevlm_results}
 
Table~\ref{tab:modality_results} compares the base model, Qwen3-VL-32B,
against DemCare-VLM, its fine-tuned counterpart, across frame counts and
question types.
 
The overall gain is substantial.
DemCare-VLM reaches 90.3 percent with a small number of frames and 93.1 percent at its best frame count, compared to the base model's peak of
83.6 percent achieved without any frames at all.
More importantly, the video-dependence gain changes sign entirely. 
For the base model it is negative, at $-3.3$ points, meaning frames hurt; for DemCare-VLM it is positive, at $+4.5$ points, meaning frames help. This is the central evidence that the fine-tuning restores genuine reliance
on the video rather than simply sharpening the same shortcut.

The per-modality breakdown complicates this picture: roughly seven-eighths of the $\Delta_v$ swing traces to the LLM-Answerable control ($\Delta_v$ moves from $-9.9$ to $-1.7$), where video grounding is irrelevant by construction, and only a small fraction to Temporal items ($+17.4$ to $+23.6$), the category the result is meant to demonstrate. This suggests the fine-tune substantially reduces the model's tendency to guess confidently when frames don't matter.


Reading the per-modality columns directly: on Temporal items, which require the ordered clip, accuracy rises by 14.5 points at $m_f{=}8$ and 13.5 points
at $m_f{=}32$, and $\Delta_v$ on this subset moves from $+17.4$ to $+23.6$. On Semantic items accuracy rises by 7.9 and 15.8 points at the two frame
counts, though $\Delta_v$ is essentially unchanged ($+53.2$ to $+52.8$), so the model is more accurate on these items without being more dependent on the
frames. On the LLM-Answerable control accuracy also rises, by 13.7 and 12.4 points, which cannot reflect improved visual reasoning since these items are
answerable from text by construction; it most likely reflects adaptation to
the vocabulary and framing of the benchmark. Because this control carries the
largest share of the split, it also carries most of the overall $\Delta_v$
swing, which is why we read the result as reduced overconfidence on
unsupported questions rather than as demonstrated temporal grounding.
 
 
The fine-tuning that produces these gains is inexpensive by comparison.
It updates a fraction of 0.12 percent of the model's parameters, using 1{,}940 training examples over three epochs, and
leaves the base weights untouched.
A lightweight adapter of this kind offers a practical route to adapting
large vision-language models to caregiving settings without the cost of
retraining the full model.

\begin{table}[t]
\centering
\caption{Overall and per-modality accuracy (\%) on the 297-question held-out
         split (the 7 paired videos excluded from DemCare-VLM training), for
         Qwen3-VL-32B (Base) versus DemCare-VLM (FT). $\Delta=\text{FT}-\text{Base}$.
         $mf$ is the number of sampled frames; $m_f{=}0$ denotes text-only
         prompting. $\Delta_v$ is the best frame-supplied accuracy minus the
         $m_f{=}0$ accuracy, computed over the frame counts shown.}
\label{tab:modality_results}
\scriptsize
\setlength{\tabcolsep}{1pt}
\begin{tabular}{@{}c rrr rrr rrr rrr@{}}
\toprule
& \multicolumn{3}{c}{\textbf{Overall}}
& \multicolumn{3}{c}{\textbf{LLM-Ans.}}
& \multicolumn{3}{c}{\textbf{Semantic}}
& \multicolumn{3}{c}{\textbf{Temporal}} \\
\cmidrule(lr){2-4}\cmidrule(lr){5-7}\cmidrule(lr){8-10}\cmidrule(lr){11-13}
$mf$ & Base & FT & $\Delta$
      & Base & FT & $\Delta$
      & Base & FT & $\Delta$
      & Base & FT & $\Delta$ \\
\midrule
0  & 83.6 & 88.6 & \gain{$+$5.0}
   & 92.2 & 96.4 & \gain{$+$4.2}
   & 20.9 & 32.6 & \gain{$+$11.7}
   & 55.3 & 62.7 & \gain{$+$7.4} \\
8  & 76.9 & 90.3 & \gain{$+$13.4}
   & 78.1 & 91.7 & \gain{$+$13.7}
   & 74.1 & 82.0 & \gain{$+$7.9}
   & 70.5 & 85.0 & \gain{$+$14.5} \\
32 & 80.4 & 93.1 & \gain{$+$12.7}
   & 82.3 & 94.7 & \gain{$+$12.4}
   & 69.6 & 85.4 & \gain{$+$15.8}
   & 72.7 & 86.3 & \gain{$+$13.5} \\
\midrule
$\Delta_v$
   & \loss{$-$3.3} & \gain{$+$4.5} & \gain{$+$7.8}
   & \loss{$-$9.9} & \loss{$-$1.7} & \gain{$+$8.2}
   & \gain{$+$53.2} & \gain{$+$52.8} & \loss{$-$0.4}
   & \gain{$+$17.4} & \gain{$+$23.6} & \gain{$+$6.2} \\
\bottomrule
\end{tabular}
\end{table}

\section{Limitations}
\paragraph{Reference-model dependence.} Modality labels come from probing a
single reference model, GPT-4o; a candidate survives only if GPT-4o answers
it correctly with ordered frames, so the benchmark contains no question
harder than GPT-4o can solve, and discarded candidates are invisible in the
released corpus. GPT-4o and Qwen3-VL-235B-A22B, which validate and generate
every question, are also evaluated on the result (Table~\ref{tab:ablation}),
so their scores should not be compared to other models on equal footing.
A panel of reference models outside the evaluation pool would remove this
dependence.
\paragraph{Actors, not patients.} The source videos are professionally
produced training material in which people with dementia are portrayed by
actors. This makes the corpus ethically distributable, but the behaviors are
dramatized: better lit, more legible, more narratively structured than real
caregiving footage. Performance here is a necessary condition for
deployment, not a sufficient one.
\paragraph{Corpus scale and coverage.} 56 videos, 94 clips. Coverage across
BPSD categories reflects the availability of training material rather than
clinical prevalence: Disinhibition contributes two videos, and three sources
account for most of the corpus. Results on the sparsest categories should be
treated as indicative.

\section{Conclusion}

DementiaCare-Bench measures a capability no existing benchmark covers: reading a dementia-care interaction in which the clinical meaning of a behaviour depends on the order and cause of events. Its central methodological claim is that a benchmark should verify, not assume, what its questions test. Applied to our own generators, modality validation shows that 77.7\% of questions written to require ordered video reduce to 34.8\% once measured, and applied to twelve current VLMs it shows that high aggregate scores rest on clinical knowledge rather than video understanding: the strongest models gain nothing measurable from frames, and several peak without them, and accuracy falls by 17 points on average on questions that require the sequence. DemCare-VLM, a LoRA adapter trained with reject-on-mismatch negatives, moves video dependence from $-3.3$ to $+4.5$ points, converting the benchmark's diagnosis into a training signal at a cost of 0.12\% of the model's parameters.

Three directions follow. First, modality validation currently depends on a single reference model; certifying labels with a panel of models outside the evaluation pool would remove the ceiling GPT-4o imposes and make labels portable across benchmarks. Second, the corpus is built from dramatized training footage, so the natural next step is validation against naturalistic recordings collected under appropriate consent, testing whether performance here transfers to the lighting, occlusion, and ambiguity of real care settings. Third, the reject-on-mismatch objective is not specific to dementia care: any VideoQA domain where language priors dominate could apply the same recipe, and establishing whether abstention-trained models remain calibrated under distribution shift is a prerequisite for deploying any of this in front of an actual caregiver.

\bibliography{refs}

\end{document}